\documentclass[lettersize,journal]{IEEEtran}
\usepackage{amsmath,amsfonts}
\usepackage{algorithmic}
\usepackage{algorithm}
\usepackage{tikz}
\usepackage{fancyhdr}
\usepackage{url}
\usepackage{booktabs}
\usepackage{balance}
\usepackage{placeins}
\AtBeginDocument{\addtolength{\abovedisplayskip}{-2pt}\addtolength{\belowdisplayskip}{-2pt}}
\usepackage{subcaption}
\usepackage{multirow}
\usepackage[font=small,labelfont=bf]{caption}
\definecolor{hkcolor}{RGB}{255,220,200}
\definecolor{pricolor}{RGB}{220,235,255}
\definecolor{enccolor}{RGB}{220,240,220}
\definecolor{deccolor}{RGB}{245,230,255}
\definecolor{rnncolor}{RGB}{255,245,210}
\definecolor{obscolor}{RGB}{235,235,235}
\definecolor{zcolor}{RGB}{255,255,255}
\definecolor{traincolor}{RGB}{30,100,180}
\definecolor{meascolor}{RGB}{180,60,30}
\definecolor{arrowcolor}{RGB}{80,80,80}
\usetikzlibrary{positioning,fit,shapes.geometric,arrows.meta,calc,decorations.pathreplacing}
\usepackage{amsmath,amssymb}
\usepackage{array}
\usepackage{tcolorbox}
\usepackage{bm}
\usepackage[font=small,labelfont=bf]{caption}

\usepackage{graphicx}
\usepackage[caption=false,font=normalsize,labelfont=sf,textfont=sf]{subfig}
 \usepackage{float}
\usepackage{textcomp}
\usepackage{caption}
\usepackage{subcaption}
\usepackage{stfloats}
\usepackage{caption}
\usepackage{caption}
\usepackage{subcaption}
\usepackage{url}
\usepackage{verbatim}
\usepackage{multicol}
\usepackage{graphicx}

\tcbuselibrary{breakable}
\usepackage{amsmath}
\usepackage{amssymb}
\usepackage[font=footnotesize,labelfont=bf]{caption}

\begin{document}

\title{Deformable Object Manipulation under Partial Observability via Real-Time Full-Shape Estimation}

\author{%
 \IEEEauthorblockN{Kosar Behnia$^1$, Ville Kyrki$^2$, Gokhan Alcan$^1$}%
 \thanks{$^1$Faculty of Engineering and Natural Sciences, Tampere Univ., Finland. }
 \thanks{$^2$Electrical Engineering and Automation Department, Aalto Univ., Finland.}
 \thanks{This work was supported by the Research Council of Finland Project under Grant 370881 and by the NVIDIA Academic Grant Program through the provision of RTX PRO 6000 Blackwell Max-Q GPUs.}%
 }

\maketitle
\thispagestyle{empty}
\begin{abstract}
Manipulating deformable objects (DOs) is challenging due to their
high-dimensional state space, underactuated dynamics, and partial
observability. In this paper, we propose cRVAE, a lightweight conditional
recurrent variational autoencoder that estimates the full DO state from only
partial corner-node observations during inference. The resulting model is used
as the forward model in a receding-horizon optimal control framework for
obstacle-aware collaborative DO manipulation. In simulation on rope and fabric, cRVAE estimates the full DO state from the
available corner-node measurements alone, matching the accuracy of a
parameter-identified XPBD model. At inference it uses no physical parameters as
model inputs and performs no online parameter identification. It also runs
$\approx 350\times$ faster on the rope and over $1500\times$ faster on the
fabric per forward pass, keeping horizon-based planning within the
$100\,\mathrm{ms}$ control budget where XPBD exceeds it already at short
horizons. Full-shape estimation from corner sensing at
in-loop speed is what makes the model deployable on hardware, which we
demonstrate on a Unitree Go2 robot. 
\end{abstract}

\section{Introduction}
Manipulating deformable objects (DOs) is a central challenge for robots operating in everyday environments, where objects such as fabrics, cables, and cleaning materials deform continuously under contact and motion.  
Unlike rigid objects, DOs have high-dimensional configurations, under-actuated dynamics and states that are only partially observable from practical sensing setups.
These properties make accurate modeling difficult, however such models are essential when a controller must predict the object shape before selecting collision-free manipulation actions.

Learning-based approaches can represent complex DO dynamics from visual, point-cloud, or dense state observations~\cite{deform,visual1,visual2,visual3}.
These methods can capture rich geometric details, including cloth wrinkles and rope curvature, and are therefore attractive for prediction and manipulation.
However, for control, such fidelity often depends on dense observations or full-state reconstruction, which can be expensive, occlusion-sensitive, and difficult to maintain on physical systems.

Physics-based models provide an alternative by propagating DO motion from a compact set of model parameters ~\cite{comparison-burak,physics-based-1,use-xpbd}.
They are attractive for model-based planning because they can be computationally efficient and interpretable.
However, when used with limited feedback, model mismatch and sim-to-real errors can accumulate over time, causing predicted shapes to drift precisely where long-horizon obstacle avoidance requires reliable forecasts.

This creates a practical gap for control-oriented DO manipulation: the planner needs a full-state, differentiable, and fast predictive model, while the physical system may provide only sparse measurements such as endpoints or corner nodes.
A suitable model should therefore infer the full DO shape from partial observations, correct its internal estimate with available measurements, and remain lightweight enough for repeated receding-horizon optimization.

\begin{figure}[t]
    \centering
    \includegraphics[width=0.85\linewidth]{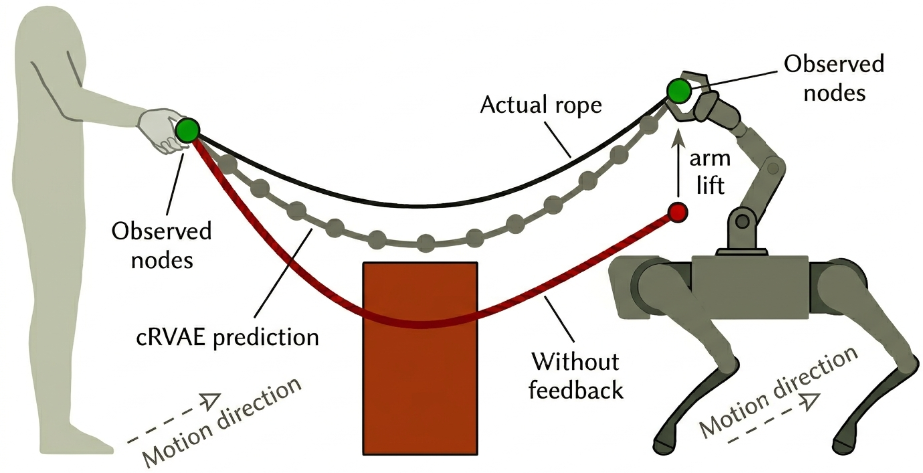}
  \caption{Collaborative rope manipulation between a human (leader) 
and a Unitree Go2 robot (follower). cRVAE estimates the full rope 
shape (grey dots) using only the two corner node measurements 
(green dots), enabling the controller to anticipate collisions and 
lift the robot arm to avoid the obstacle (black rope). Without a 
full-shape estimate, the controller cannot anticipate the 
collision and the rope hits the obstacle (red rope).}
    \label{fig:teaser}
\end{figure}
\par In this work, we propose cRVAE, a control-oriented conditional recurrent variational autoencoder that estimates the full DO state from sparse corner-node measurements and manipulation inputs.

During inference, cRVAE replaces the encoder with a learned conditional prior and injects measurement feedback into both the estimated state and the recurrent hidden state, reducing estimation error when measurements arrive at a high rate.

The resulting model is differentiable and lightweight enough for receding-horizon optimization, while retaining the full-shape predictions needed for obstacle-aware manipulation, as illustrated in Fig.~\ref{fig:teaser}. 
The main contributions of this work are as follows:

\begin{itemize}
    \item We introduce cRVAE, a conditional recurrent VAE for full-state DO estimation from sparse corner-node observations.
    \item We integrate the learned differentiable model into a receding-horizon optimal control framework for collaborative, obstacle-aware DO manipulation under partial observability.
    \item We show that cRVAE estimates the full state from the available corner-node observations and attains accuracy comparable to a parameter-identified XPBD model without physical parameters as model inputs or online identification during inference, while requiring $\approx 9$\,ms per rollout on the rope against $3.2$\,s ($\approx 360\times$, both including controller derivatives) for XPBD, meeting the real-time budget required for in-loop use and enabling a horizon-based controller to complete narrow-gap tasks where a reactive baseline fails.
\end{itemize}

\section{Related Works}

A model that can accurately predict the future full state of 
a deformable object is needed to manipulate it with a robot. 
Such a model has to be fast enough to be used inside an 
optimization-based controller, such as iLQR or MPC, and 
remain accurate despite the partial observation and limited 
sensing typically available on a physical system. We organize 
related work around these three requirements: full-state 
prediction models for DOs, predictive models for 
optimization-based DO control, and methods for handling 
partial observation and feedback in real-world DO manipulation.

a) Full-State Prediction Models for DOs: Learning-based 
models have been widely used to obtain compact 
representations of DO dynamics from high-dimensional 
observations. PointNet-based encoders, for example, can map 
RGB-D point clouds to a compact latent space 
\cite{empirical}, while variational autoencoders map dense 
RGB observations into a latent space in which the dynamics 
are modeled by a learned function of the previous latent 
state \cite{yan2020contrastive}. This function can take the form of an MLP 
\cite{deform}, a linear model \cite{linear, visual2}, an RBF 
\cite{yu2023global}, a recurrent state-space model 
\cite{ropedreamer2026}, or a 
graph-based model that captures relations between object 
parts \cite{visual1}. These methods can capture rich 
geometric detail and generalize across tasks, but they 
typically require dense or full-state observations during 
training and inference, which can be costly and difficult to 
obtain on physical systems.
Physics-based models offer an alternative, using a compact 
set of physical parameters, such as stiffness and mass, to 
propagate DO motion directly, as in position-based dynamics 
\cite{XPBD_Macklin}. Such models have been used for 
collaborative obstacle avoidance \cite{comparison-burak}, for 
dual-arm cable manipulation \cite{zhu2018iros}, and 
for planning and control of deformable linear objects 
\cite{physics-based-1}. These models are interpretable  once their parameters are known, 
but their accuracy depends on simulator parameters, such as 
stiffness and damping coefficients, that must be carefully 
identified to match the real object. Overall, many existing 
full-state predictors need dense sensing, full-state 
reconstruction, or accurate simulator parameters, which are 
difficult to obtain in practical manipulation settings. In 
contrast, our proposed method, cRVAE, estimates the full DO 
state using only sparse corner-node observations, without 
requiring full DO node observations or carefully identified 
simulator parameters at inference.

b) Predictive Models for Optimization-Based DO Control: To 
use optimization-based control strategies such as MPC, a 
model of the DO dynamics is required that is both 
differentiable and fast enough to evaluate repeatedly at 
every planning step. Several works learn this model directly. 
For example, \cite{differentiable-shapecontrol-fullnodes} 
learns a model of DLO shape that can be differentiated 
through within gradient-based MPC, while 
\cite{diff-whole-nodes-2} learns a state representation using 
a differentiable physics engine. However, both approaches 
require full DO node observations during inference, which can 
be computationally expensive and difficult to obtain in 
practice. Other approaches use physics-based simulators, such 
as XPBD \cite{use-xpbd, comparison-burak} or PBD 
\cite{physics-based-1}, to obtain derivatives directly for use 
within a differentiable controller for DO manipulation. 
However, computing gradients through such physics-based 
simulators can be computationally expensive, and their 
accuracy remains sensitive to mismatch between the simulated 
and real object parameters. Overall, many of these methods are 
useful for planning but can be costly, require dense state, or 
are sensitive to model mismatch. In contrast, cRVAE is 
differentiable and provides fast gradient computation.

c) Partial Observation and Feedback in Real-World DO 
Manipulation: Deploying DO manipulation methods in practice 
is challenging because predictions made in simulation do not 
always transfer to the real world, especially under limited 
sensing  \cite{blancomulero2024}. Physics-based models, for instance, can fail in 
practice without real measurement feedback, as model mismatch 
and sim-to-real error accumulate over time \cite{physics-based-1}. 
To address this, some methods calibrate simulator parameters 
online using a residual mapping between the simulation and 
real observations \cite{liang2024}, but each update of this 
online optimization can take up to several seconds, limiting 
its use within a real-time receding-horizon controller. Rather 
than calibrating a full-state model, other methods avoid 
predicting the full state altogether. \cite{toner2024} 
controls only the endpoint poses of a deformable linear 
object, without estimating its full pose. \cite{sirintuna2024} 
does not estimate the DO's shape for human-robot collaborative 
transportation, and \cite{zhang2026} treats the DO as 
effectively rigid during formation planning for a similar 
task. Existing sparse-feedback methods often do not 
reconstruct or control the full DO shape over long horizons, 
instead relying on local corrections or task-specific, 
output-based formulations. In contrast, cRVAE reconstructs 
the full DO shape from sparse measurements, without an 
iterative online optimization step, making it suitable for 
real-time receding-horizon control.

\section{Problem Formulation}
\label{sec:system_formulation}

We model a deformable object as a set of $N$ nodes. The position of the $i$-th node at time step $k$ is denoted by $x^i_k \in \mathbb{R}^3$, and the full object state is represented by the stacked vector
\begin{equation}
    P_k = [(x^1_k)^\top,\ldots,(x^N_k)^\top]^\top \in \mathbb{R}^{3N},
    \label{eq:full_state}
\end{equation}

Given a manipulation input $f_k \in \mathbb{R}^m$, the deformable object dynamics are modeled by an unknown transition function
\begin{equation}
    P_{k+1} = F(P_k, f_k).
    \label{eq:dynamics}
\end{equation}
In practical manipulation tasks, the full state $P_k$ is not directly available during inference. Instead, the robot observes only a sparse subset of $K<N$ boundary or corner nodes, indexed by $\mathcal{I}=\{c_1,\ldots,c_K\}$,
\begin{equation}
    \mathcal{C}_k = \{x^{c_1}_k, \ldots, x^{c_K}_k\} 
    \in \mathbb{R}^{3K}.
    \label{eq:partial_obs}
\end{equation}
The generic problem considered in this paper is to obtain a control-oriented predictor that estimates the full future object state from sparse observations and manipulation inputs. More generally, the predictor may maintain an internal state $\eta_k$ that summarizes past observations and predictions,

\begin{equation}
    (\hat{P}_{k+1},\eta_{k+1})
    = \Phi_\theta(\mathcal{C}_k, f_k, \eta_k),
    \label{eq:generic_predictor}
\end{equation}
where $\hat{P}_{k+1}$ is the estimated full DO state, and we denote by
$\widehat{\mathcal{C}}_k$ the entries of $\hat{P}_k$ at the measured node
indices $\mathcal{I}$, i.e.\ the predicted counterpart of
$\mathcal{C}_k$ in~\eqref{eq:partial_obs}. The desired predictor should be accurate enough for obstacle-aware planning and computationally light enough to be evaluated repeatedly inside a receding-horizon controller.

\section{Proposed Method}

The formulation in Section~\ref{sec:system_formulation} requires a predictor that can infer the full DO state from sparse measurements, retain temporal information, and remain differentiable and lightweight for receding-horizon optimization. To instantiate the generic predictor $\Phi_\theta$ in~\eqref{eq:generic_predictor}, we propose cRVAE, a conditional recurrent variational autoencoder. 

\begin{figure}[t]
    \centering
    \includegraphics[width=0.9\linewidth]{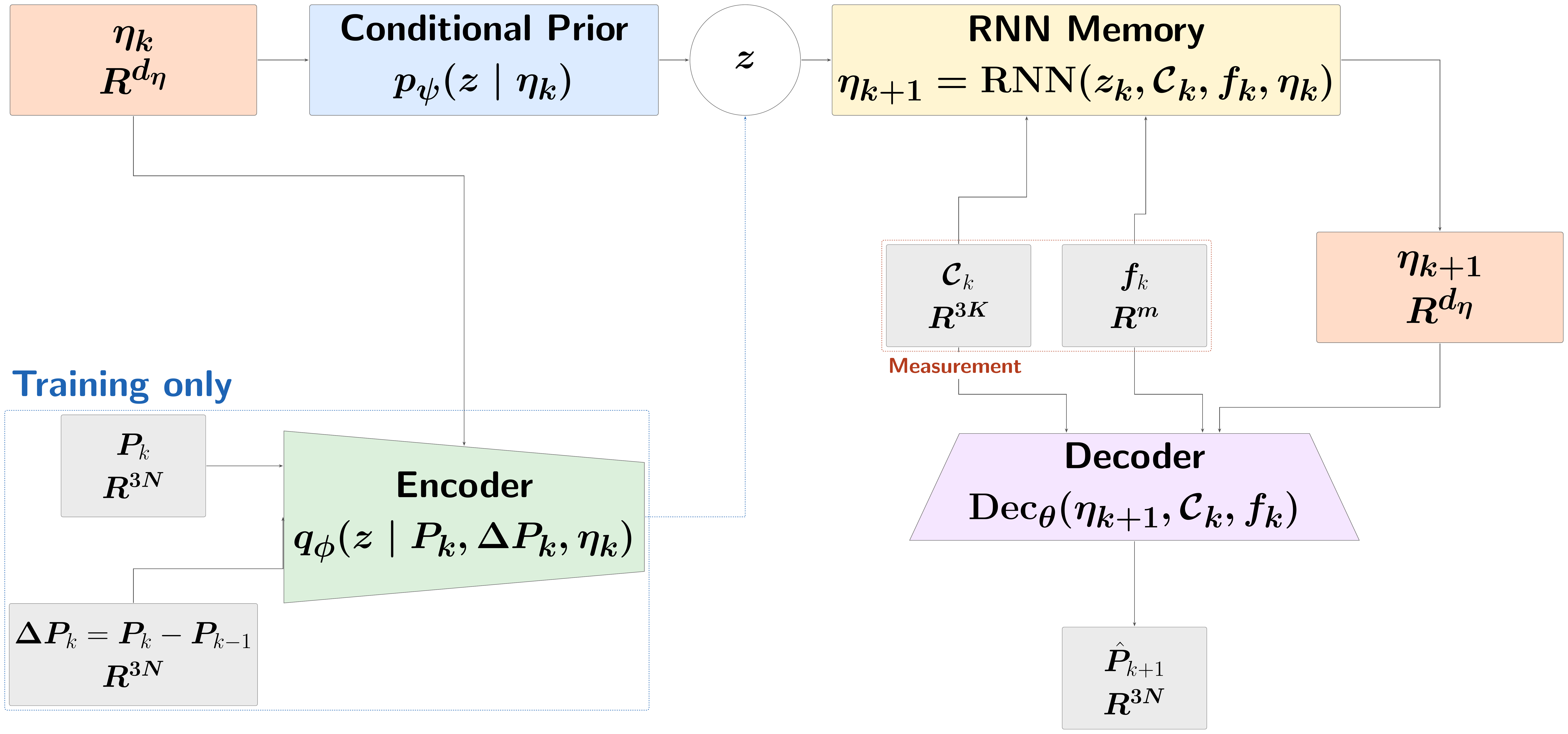}
    \caption{Architecture of the proposed cRVAE model. During training, the encoder posterior  $q_\phi(z \mid P_k, \Delta P_k, \eta_k)$ is used to sample the latent variable $z$; during inference, it is replaced by the conditional prior $p_\psi(z \mid \eta_k)$.
    }
    \label{fig:crvae_architecture}
\end{figure}

\textit{During training}, full simulated states are available and are used to learn a posterior latent distribution and reconstruct the next full state. \textit{During inference}, the full state is unavailable, therefore, the encoder is replaced by a learned conditional prior and the decoder estimates the full state using only the hidden state, sparse measurements, and manipulation input. Measurement feedback from the available nodes is then used to correct both the estimated state and the recurrent hidden state, limiting drift during multi-step prediction. 
Fig.~\ref{fig:crvae_architecture} shows the overall architecture of cRVAE. 

In the following subsections, we detail each model component (Section~\ref{sec:model_components}), the inference procedure with measurement feedback (Section~\ref{sec:measurement_feedback}), and the control problem formulation (Section~\ref{sec:control}).

\subsection{Model Components}
\label{sec:model_components}

\subsubsection*{a) Encoder}
The encoder maps the current full state $P_k$, its temporal difference $\Delta P_k = P_k - P_{k-1}$, and the previous hidden state $\eta_k$ to the parameters of a Gaussian posterior:
\begin{equation}
    \mu_k,\, \log\sigma^2_k = \text{Enc}_\phi(P_k, \Delta P_k, \eta_k),
    \quad z_k \sim \mathcal{N}(\mu_k, \sigma^2_k).
    \label{eq:encoder}
\end{equation}

\subsubsection*{b) Conditional Prior}
The prior estimates the latent distribution conditioned only on the hidden state:
\begin{equation}
    \mu^p_k,\, \log(\sigma^p_k)^2 = \text{Prior}_\psi(\eta_k),
    \quad z_k \sim \mathcal{N}(\mu^p_k, (\sigma^p_k)^2).
    \label{eq:prior}
\end{equation}

\subsubsection*{c) Decoder}
The decoder reconstructs the full next state from the hidden state, and the partial observations:
\begin{equation}
    \hat{P}_{k+1} = \text{Dec}_\theta( \eta_{k + 1}, \mathcal{C}_k \textcolor{black}{,f_k}).
    \label{eq:decoder}
\end{equation}

\subsubsection*{d) RNN Module}
The hidden state is updated at each step as:
\begin{equation}
  \eta_{k+1} = \text{RNN}(z_k, \mathcal{C}_k, f_k, \eta_k).
    \label{eq:rnn}
\end{equation}

\subsubsection*{e) Training Objective}
The model is trained by minimizing a combination of reconstruction loss and KL divergence:
\begin{eqnarray}
\mathcal{L} &=& \sum_{k=1}^{T-1} \left[
\underbrace{\| P_{k+1} - \hat{P}_{k+1} \|^2}_{\text{Reconstruction}}
+  \underbrace{w_c \| \mathcal{C}_{k+1} - \widehat{\mathcal{C}}_{k+1} \|^2}_{\text{Corner nodes}}
\right.
\label{eq:loss}
\\ \nonumber
&+& \left.
\lambda
\underbrace{
D_{\text{KL}}\left(
q_\phi(z_k \mid P_k,\Delta P_k,\eta_k)
\,\|\, p_\psi(z_k \mid \eta_k)
\right)
}_{\text{KL Divergence}}
\right]
\end{eqnarray}
where $\lambda$ balances reconstruction accuracy against the consistency
between the encoder posterior and the conditional prior, and $w_c$ weights the
corner-node term. 
\subsection{Inference with Measurement Feedback}
\label{sec:measurement_feedback}
During inference, the estimated corner node positions $\widehat{\mathcal{C}}_k$ may drift from the real sensor measurements due to model errors and the sim-to-real gap. To address this, 
we define the residual $r_k \in \mathbb{R}^{3K}$ between the measured corner nodes and the corresponding estimated values as:
\begin{equation}
    r_k = \mathcal{C}^{\mathrm{dyn}}_k - \widehat{\mathcal{C}}_k,
    \label{eq:residual}
\end{equation}
where $\mathcal{C}^{\mathrm{dyn}}_k$ is the corner node position, obtained from the actual measurement at every $T_m$ steps and propagated by the input dynamics in between:
\begin{equation}
    \mathcal{C}^{\mathrm{dyn}}_k = \begin{cases} 
    \mathcal{C}_k & \text{if } k \mod T_m = 0, \\
    g(\mathcal{C}^{\mathrm{dyn}}_{k-1}, f_{k-1}) & \text{otherwise,}
    \end{cases}
    \label{eq:ydyn}
\end{equation}
where $g(\cdot)$ is a known propagation function that estimates the corner node positions between measurement steps, which can be obtained, for example, from the robot kinematics or dynamics. In this work, we consider the case where $f_k$ represents the velocity of the actuated node, giving $g(\mathcal{C}^{\mathrm{dyn}}_{k-1}, f_{k-1}) = \mathcal{C}^{\mathrm{dyn}}_{k-1} + \Delta t\, f_{k-1}$.

The residual $r_k$ is used to correct both the estimated full 
state and the hidden state via:
\begin{align}
\hat{P}^{\text{corr}}_k &= \hat{P}_k + \alpha\, J^{\text{nodes}}_k\, r_k, \label{eq:state_correction} \\
    \eta^{\text{corr}}_k &= \eta_k + \beta\, J_k^\top\, r_k,
    \label{eq:hidden_correction}
\end{align}
where $J^{\text{nodes}}_k = \frac{\partial \hat{P}_k}{\partial \mathcal{C}_k} 
\in \mathbb{R}^{3N \times 3K}$ and $J_k = \frac{\partial \widehat{\mathcal{C}}_k}{\partial \eta_k} \in \mathbb{R}^{3K \times d_{\eta}}$ are the Jacobians 
of the estimated state with respect to the corner node 
positions and the internal state, respectively, and $\alpha$, 
$\beta$ are gains controlling the trust between the model 
estimate and the measurement.

By correcting the hidden state directly, the feedback ensures that the temporal context carried by the RNN is also adjusted, limiting error accumulation and partially compensating for the sim-to-real gap. The corrected state $\hat{P}^{\text{corr}}_k$ is not fed back into
the recurrence, which is anchored through $\eta^{\text{corr}}_k$
alone; it is the measurement-corrected estimate of the current
state, and is used to initialize the receding-horizon problem of
Section~\ref{sec:control} at each planning cycle.

\subsection{Control Problem Formulation}
\label{sec:control}

Given the learned model $\Phi_\theta$, we formulate the 
DO manipulation task as a receding-horizon optimal control 
problem. At each planning cycle, the following problem is 
solved over a horizon of $H$ steps:

\begin{align*}
   \text{Given:} \quad & \mathcal{O} \subset \mathbb{R}^3,\ \mathcal{C}_0,\ \eta_0, \\[2pt]
    \min_{f_0, \ldots, f_{H-1}} \quad &
    \mathcal{J} = \sum_{k=0}^{H-1} \ell(\hat{P}_k, f_k)
        + \ell_T(\hat{P}_H) \\[2pt]
    \text{subject to:} \quad &
    \mathcal{D}(\hat{P}_k, f_k, \mathcal{O}) \geq 0, \\ 
& (\hat{P}_{k+1}, \eta_{k+1}) = \Phi_\theta(\mathcal{C}_k, f_k, \eta_k),\end{align*}
where $f_k \in \mathbb{R}^m$ is the applied input to be 
optimized, $\ell(\cdot)$ and $\ell_T(\cdot)$ are the running 
and terminal cost functions that can encode, for example, a 
tracking error toward a desired DO configuration, and 
$\mathcal{D}(\cdot)$ represents the inequality constraint 
function defining the feasible region $\mathcal{O} \subset 
\mathbb{R}^3$. The dynamics constraint uses $\Phi_\theta$ as the forward model, with the
corner node inputs propagated within the horizon by~\eqref{eq:ydyn}.

\section{Experiments}

The proposed method is evaluated in three stages: (1) a rope simulation, assessing cRVAE's prediction accuracy and its use in leader-follower obstacle avoidance (Section~\ref{sec:rope_sim}); (2) a fabric simulation, extending the evaluation to a 2D deformable object (Section~\ref{sec:fabric_sim}); (3) real-world rope manipulation on a Unitree Go2 robot, assessing the framework on physical hardware under the sim-to-real and real-to-sim gap (Section~\ref{sec:realworld_exp}).
The same cRVAE architecture is used throughout: encoder 512--256, prior
256, decoder 256--512 (ReLU MLPs), and a GRU with $d_{\eta}{=}32$
and latent dimension 32, giving 491k parameters for the rope
($N{=}33$, $K{=}2$, $m{=}3$) and 1.38M for the fabric ($N{=}225$, $K{=}4$,
$m{=}6$), of which 225k and 524k remain at inference once the encoder is
discarded. Both are trained with Adam (lr $10^{-3}$, batch 16), $\lambda$
annealed $0\!\to\!0.05$ over 40 epochs, and a rollout curriculum from 1 to 10 steps.
\subsection{Rope Simulation}
\label{sec:rope_sim}
\subsubsection*{a) cRVAE Training and Modeling Accuracy}
The cRVAE is trained on rope dynamics generated by an XPBD-based simulator \cite{comparison-burak}, where the rope, with a length of $1.5$\,m, is discretized into $N = 33$ nodes. Node 27 is selected as the actuated node, with the applied input $f(k)$ defined as its velocity, using a time step of $\Delta t = 0.1$\,s. Node 3 is taken as the other corner node, so the corner node measurement is given by $\mathcal{C}_k = \{x^{27}_k, x^{3}_k\}$. The training dataset consists of 10 trajectories, each approximately one minute long, covering circular movements, translations along the $x$, $y$, and $z$ directions, and random motions. Fig.~\ref{fig:mse_horizon}  shows the cRVAE prediction error versus horizon, reported as mean absolute error (MAE) per coordinate, averaged over all nodes, axes, and time steps (cm). We compare against two references. The geometric baseline holds the rope's initial shape and maps it at each step by the least-squares similarity transform (rotation, scale, translation) aligning its two corner nodes to the observed corners. The physics reference is an XPBD rope model with stretching and bending
constraints on point particles, with parameters identified on a held-out
sequence. The generator \cite{comparison-burak} additionally models orientation and bend--twist constraints, so the reference cannot represent torsion and part of its error is structural rather than a consequence of poor identification. As shown in Fig.~\ref{fig:mse_horizon}, cRVAE reaches $0.81$~cm at $h=1$ and $2.51$~cm at $h=40$, against $2.75$ and $3.28$~cm for the physics reference and $2.79$ and $5.66$~cm for the baseline. The physics reference starts near the baseline at $h=1$ but grows more slowly, ending $2.38$~cm below it at $h=40$, while cRVAE stays below both across the horizon.
\subsubsection*{b) Ablation of the Latent Variable}
\begin{figure}[b!]
    \centering
    \includegraphics[width=0.9\linewidth]{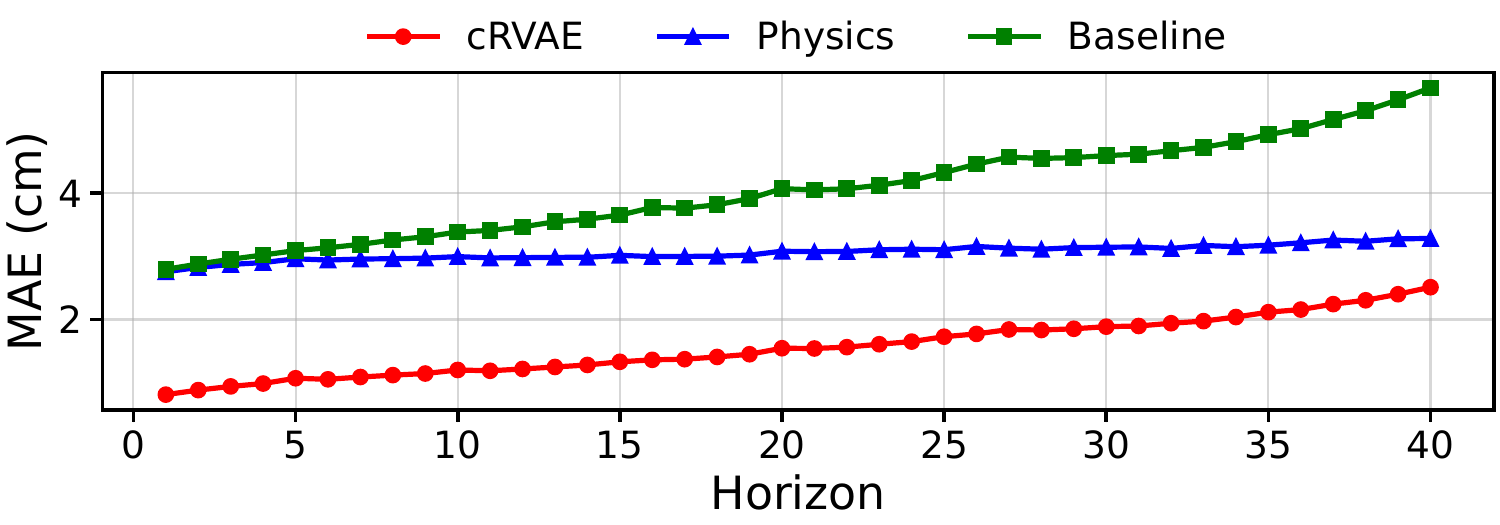}
  \caption{cRVAE prediction MAE (cm) versus prediction horizon on the rope simulation, against a similarity-transform baseline and an XPBD rope reference.}
    \label{fig:mse_horizon}
\end{figure}

To isolate the effect of the stochastic latent prior, we retrain a
deterministic variant in which the prior is replaced by a feature map of
identical width and depth, with the encoder and KL term removed and all
else unchanged (parameter counts within $4\%$). Both are retrained and
evaluated under the same protocol, but on a different held-out set from
Fig.~\ref{fig:mse_horizon}, so the values are comparable to each other but not to that figure. cRVAE is more accurate at every horizon tested, by
$0.57$~cm at $h=1$ and $0.55$~cm at $h=10$, indicating that the latent
variable contributes beyond the recurrent structure alone.

\subsubsection*{c) What Do the Corner Measurements Anchor?}
We ablate the feedback correction
(\eqref{eq:state_correction}--\eqref{eq:hidden_correction}) on nine held-out
rope trajectories, comparing no feedback ($\alpha{=}\beta{=}0$), $\alpha$ only,
$\beta$ only, and full feedback across measurement intervals $T_m$ (as
in~\eqref{eq:ydyn}, distinct from the prediction horizon $h$). The gains
were selected by a grid search on a single training trajectory
($\alpha \in \{0,0.3,0.5,1.0\}$, $\beta \in \{0,1,2,3\}$) minimizing
interior-node MAE, giving $\alpha{=}1.0$, $\beta{=}2.0$, held fixed for all
evaluations. Since the
correction is derived entirely from the two measured corners, the unmeasured
interior nodes are the meaningful test. As every condition is evaluated on the
same trajectories, we test the paired per-trajectory differences and report the
Wilcoxon signed-rank $p$ alongside the number of trajectories improved.

At short measurement intervals the correction is effective and consistent in
direction: the full correction reduces interior MAE from $1.256$ to
$1.058$~cm at $T_m{=}1$ ($+13.1\% \pm 7.2$) and from $1.226$ to $1.058$~cm at
$T_m{=}5$ ($+10.4\% \pm 8.7$), improving all nine held-out trajectories in both
cases ($p = 0.004$). The measured corners improve considerably more
($+65.4\%$ at $T_m{=}1$), as expected since the residual constrains them
directly. The per-trajectory spread is comparable to the mean, reflecting
genuine variation in how much each motion benefits; the direction of the
effect, however, is consistent across every trajectory. The benefit decreases as the measurement interval grows, and by $T_m{=}10$ no
condition is statistically distinguishable from the no-feedback baseline
($p \ge 0.16$). To understand why the correction saturates, we ask how much interior shape each measurement set identifies: at each step we fit the hidden state to reproduce the measured nodes (by gradient descent through the frozen decoder) and score the unmeasured interior. Fitting the full state reaches $0.35$~cm ($+72\%$ over no feedback), so a good interior is representable by the decoder, whereas fitting only the two corners is net-negative ($2.04$~cm, $-62\%$): many hidden states reproduce them equally well with different interiors, so an unconstrained fit drifts, while the damped update~\eqref{eq:state_correction} stays near an already-plausible prediction. Adding node~12 reaches $0.56$~cm ($+56\%$). The corners thus carry little interior information, which is why the $\alpha$--$\beta$ gain is small. Such an interior measurement can enter the hidden-state correction~\eqref{eq:hidden_correction} without appearing among the network inputs, so no retraining is needed: replacing the fixed-gain update with one damped gradient step on $E(\eta)=\|\hat{P}(\eta)|_{\{3,12,27\}}-\mathcal{C}^{+}\|^2 $, where $\mathcal{C}^{+}$ denotes the two corners augmented with node~12, reduces interior MAE, scored on the nodes measured by neither the corners nor the added sensor, from $1.29$ to $0.88$~cm at $T_m{=}1$ ($+28\%$, $p=0.004$, 9/9). Sensor placement, rather than gain tuning, is therefore the lever that matters.

\subsubsection*{d) Comparison}
The proposed method is compared against \cite{comparison-burak} 
in a rope manipulation task where node 3 acts as the leader 
and node 27 as the follower, which should maintain a desired 
offset of $(0, -0.4, -0.5)$\,m in the $x$, $y$, and $z$ 
directions, respectively, while avoiding obstacles.
The leader moves at mean speeds of $0.005$, $0.015$, $0.025$ and $0.05$\,m/s in the Baseline scenario, and at $0.005$ and $0.015$\,m/s in the Challenging scenario, where running XPBD in the loop is too costly to repeat at every speed. The standard deviation is $10\%$ of the mean and each setting is repeated 10 times.
In some scenarios, such as the Baseline scenario 
described below, both methods succeed regardless of 
horizon-based planning. However, in other configurations, such 
as the Challenging scenario, the feasible path between 
obstacles is narrow enough that a horizon-based controller is 
needed to find a collision-free trajectory. We \textcolor{black}{use 
iLQR~\cite{ilqr}} as the horizon-based controller. Since iLQR requires the rope's state at each future step within the horizon, cRVAE is used to predict these intermediate states. 

A run is considered successful if the smoothed minimum distance to the obstacle never remains below the safety threshold for longer than $t > 0.5$\,s, distinguishing real contact from transient measurement noise. In addition to the success criterion, we report the clearance, defined as the smallest value of the smoothed distance to the obstacle observed during a run, with its mean and standard deviation reported across all runs in a scenario.

\textit{Baseline Scenario:} Two rectangular obstacles are placed
at $x=0$, centered at $(y,z)=(\pm 0.85, 0.75)$~m, each $1.0$~m
wide and $1.5$~m high.  Both methods complete all runs at all four speeds with positive clearance, confirming that neither collides with the obstacle. cRVAE's clearance falls with speed ($0.15$, $0.09$, $0.05$, $0.04$~m), while CBF-QP \cite{comparison-burak} stays near $0.04$--$0.05$~m up to $0.025$\,m/s and reaches $0.12$~m at $0.05$\,m/s. At the highest speed CBF-QP therefore keeps the larger margin, while cRVAE stays closer to the nominal shape and trades clearance for tracking accuracy.

\begin{table}[t!]
    \centering
    \vspace*{5pt}
    \caption{Task Completion -- Baseline Scenario. Both methods complete
    $100\%$ of runs at every speed.}
    \label{tab:safety_baseline}
    \resizebox{\linewidth}{!}{
    \begin{tabular}{lcccc}
        \hline
        \textbf{Clearance (m)} & $0.005$ & $0.015$ & $0.025$ & $0.05$ \\
        \hline
        cRVAE (Ours) & $0.15 \pm 0.003$ & $0.09 \pm 0.004$ & $0.05 \pm 0.01$ & $0.04 \pm 0.003$ \\
        CBF-QP \cite{comparison-burak} & $0.05 \pm 0.004$ & $0.04 \pm 0.01$ & $0.04 \pm 0.01$ & $0.12 \pm 0.016$ \\
        \hline
    \end{tabular}}
    \vspace{0.5pt}
    \footnotesize{Columns are leader speeds in m/s.}
\end{table}

\textit{Challenging Scenario:} The obstacle configuration is modified by placing two obstacles vertically stacked at $x = 0$. The lower obstacle has center $(y, z) = (0.75, 0.75)$\,m, width $0.8$\,m and height $1.5$\,m, and the upper obstacle has center $(y, z) = (0.75, 2.95)$\,m with the same dimensions. 

\begin{figure}[t!]
    \centering
    \begin{minipage}[t]{0.32\linewidth}
        \centering
        \includegraphics[width=\linewidth]{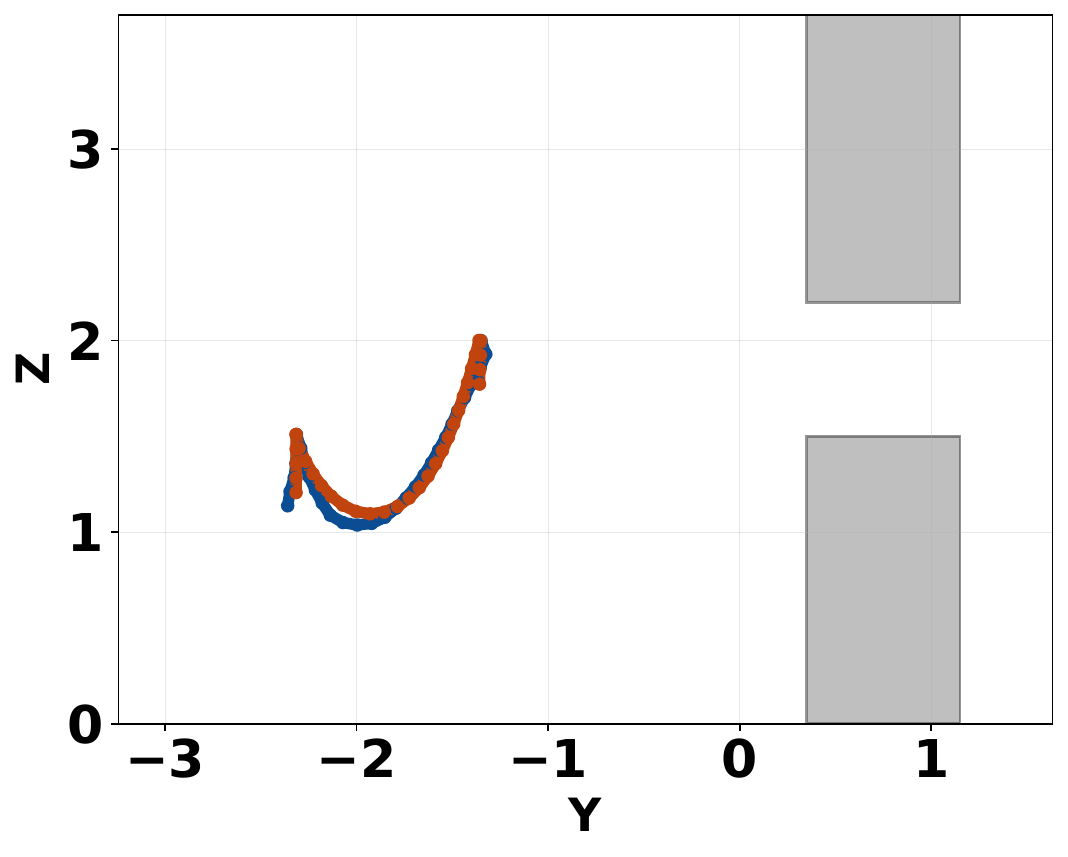}
        \par\smallskip
        {\small (a) $t = 0.0$ s.}
        \label{fig:rope_stretch_before}
    \end{minipage}
    \hfill
    \begin{minipage}[t]{0.32\linewidth}
        \centering
        \includegraphics[width=\linewidth]{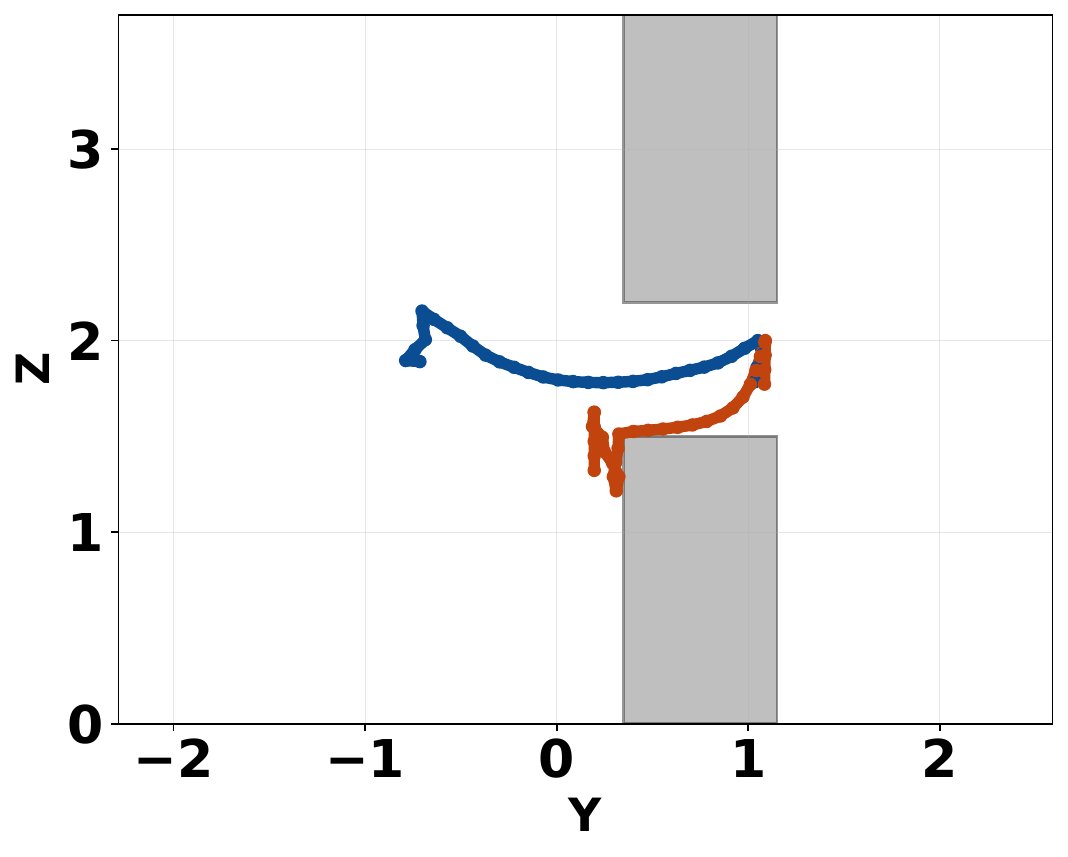}
        \par\smallskip
        {\small (b) $t = 5.2$ s.}
        \label{fig:rope_stretch_at}
    \end{minipage}
    \hfill
    \begin{minipage}[t]{0.32\linewidth}
        \centering
        \includegraphics[width=\linewidth]{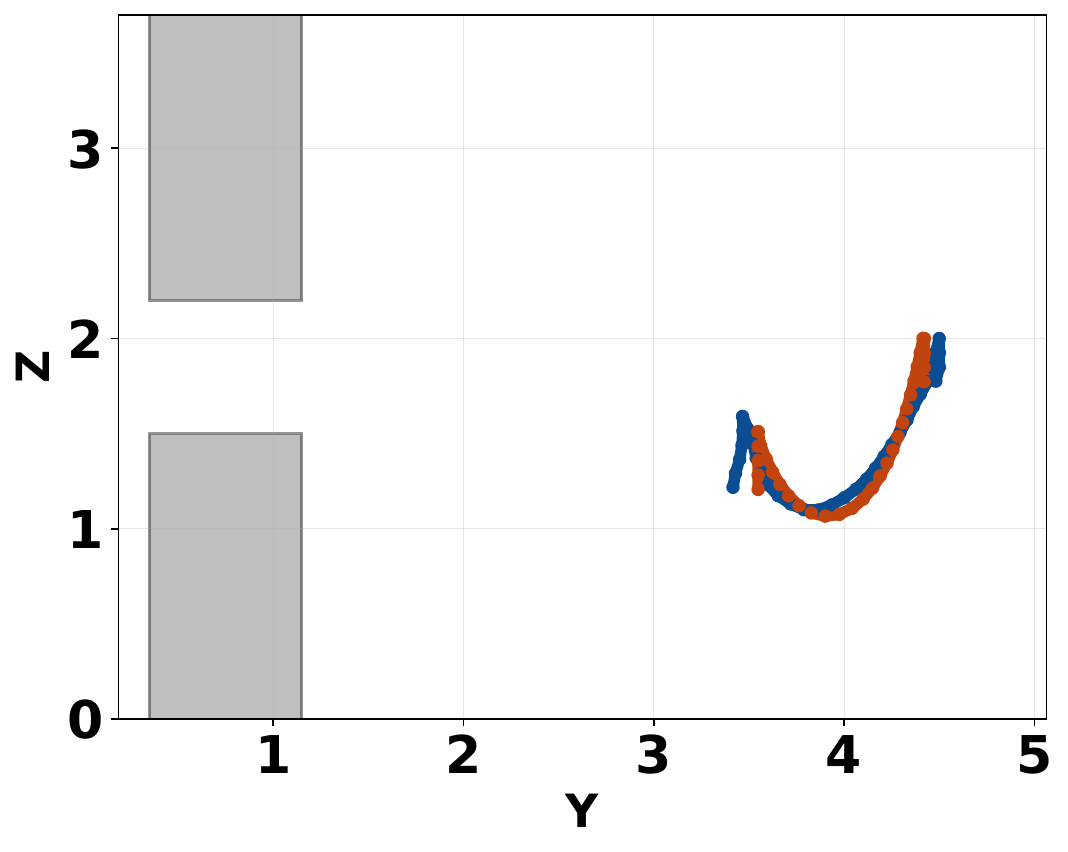}
        \par\smallskip
        {\small (c) $t = 12.3$ s.}
        \label{fig:rope_stretch_after}
    \end{minipage}
  \caption{Rope shape near the obstacle in the Challenging
scenario, shown at $t = 0.0$\,s before reaching the obstacle,
$t = 5.2$\,s at the obstacle, and $t = 12.3$\,s after passing
the obstacle.}
    \label{fig:rope_stretch}
\end{figure}

\begin{figure}[t!]
    \centering
    \includegraphics[width=0.9\linewidth]{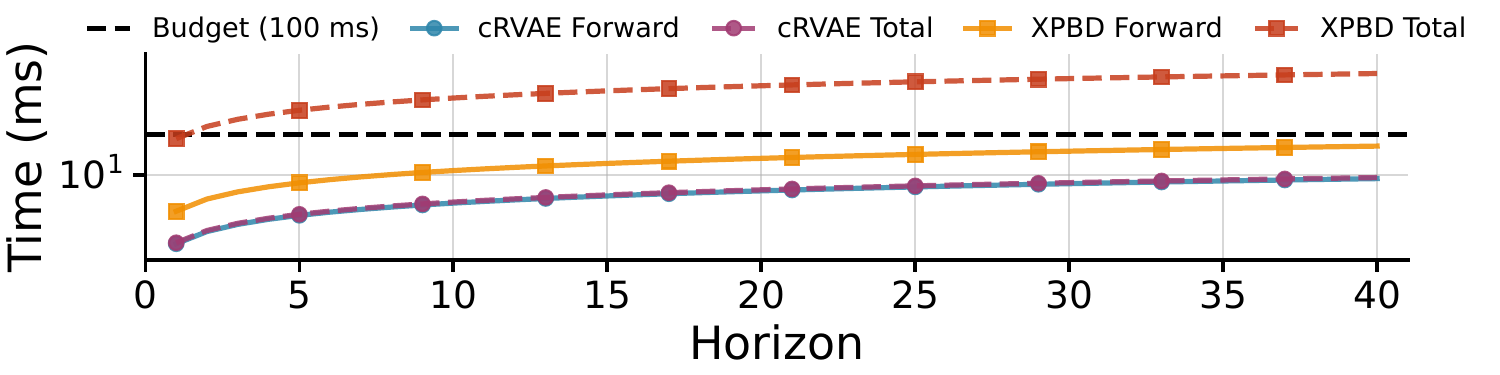}
  \caption{Per-horizon compute time on the rope (log scale) versus horizon $h$, 
for cRVAE and XPBD. ``Total'' includes forward prediction and derivative 
computation using each model.}
    \label{fig:timing}
\end{figure}

As reported in Table~\ref{tab:safety_challenging}, the proposed method completes the task successfully across all speed scenarios, maintaining a consistent clearance of approximately $0.10$\,m regardless of speed, while \cite{comparison-burak} fails to complete the task and reaches zero clearance. Passing through a narrow gap requires the rope to stretch, which is achieved by planning the deformation ahead using the horizon-based controller. Fig.~\ref{fig:rope_stretch} shows the rope shape at $t = 0.0$\,s before reaching the obstacle, $t = 5.2$\,s at the obstacle, and $t = 12.3$\,s after passing the obstacle. At $t = 5.2$\,s, the proposed method stretches the rope along the $y$-axis to clear the gap, while \cite{comparison-burak} collides with the lower obstacle. At $t = 0.0$\,s and $t = 12.3$\,s, the rope shapes of the two methods are near each other. 
Since \cite{comparison-burak} is a reactive CBF-QP controller, the contrast in
Table~\ref{tab:safety_challenging} reflects the benefit of horizon-based
planning rather than a difference between the two models. To isolate the
forward model from the controller, we additionally run XPBD as the forward
model inside the same iLQR loop, under identical cost, constraints, initial
conditions, and plant, differing only in the model used for prediction and its
derivatives. As reported in Table~\ref{tab:safety_challenging}, XPBD+iLQR also
completes the task and avoids the obstacle, with clearances comparable to
cRVAE across both speeds.
Under the same horizon-based controller, cRVAE thus achieves task performance
comparable to XPBD while reconstructing the full rope state from two observed
nodes, without parameter identification at inference, and with substantially faster
forward and derivative computation.
 As shown in Fig.~\ref{fig:timing}, XPBD
exceeds the $100$\,ms control budget at every horizon, 
taking $396$\,ms at
$h=5$ and $3.2$\,s at $h=40$, whereas cRVAE stays well within budget
($\approx 9$\,ms at $h=40$);
both are total times, including the derivatives
iLQR requires. For a fair comparison, XPBD is reimplemented in JAX so that both models are
JIT-compiled, differentiated by the same autodiff system, and timed on the same
GPU, with each model's forward pass and iLQR Jacobians measured separately. XPBD is therefore accurate enough to plan with
but too slow to deploy in the loop, which is precisely the gap cRVAE closes.

\begin{table}[t!]
    \centering
    \caption{Task Completion -- Challenging Scenario}
    \label{tab:safety_challenging}
    \resizebox{\linewidth}{!}{
  \begin{tabular}{lccc}
    \hline
    \textbf{Method}
    & \textbf{Task Completion (\%)}
    & \textbf{Leader Speed (m/s)}
    & \textbf{Clearance (m)} \\
    \hline
    cRVAE (Ours)              & $100\%$ & $0.005$ & $0.097 \pm 0.014$ \\
    XPBD + iLQR               & $100\%$ & $0.005$ & $0.127 \pm 0.039$ \\
    CBF-QP \cite{comparison-burak} & $0\%$ & $0.005$ & $0.0 \pm 0.0$ \\
    \hline
    cRVAE (Ours)              & $100\%$ & $0.015$ & $0.098 \pm 0.029$ \\
    XPBD + iLQR               & $100\%$ & $0.015$ & $0.073 \pm 0.044$ \\
    CBF-QP \cite{comparison-burak} & $0\%$ & $0.015$ & $0.0 \pm 0.0$ \\
    \hline
\end{tabular}}
\end{table}


\subsection{Fabric Simulation}
\label{sec:fabric_sim}
\subsubsection*{a) cRVAE Training and Modeling Accuracy}
\begin{figure}[b!]
    \centering
    \includegraphics[width=0.9\linewidth]{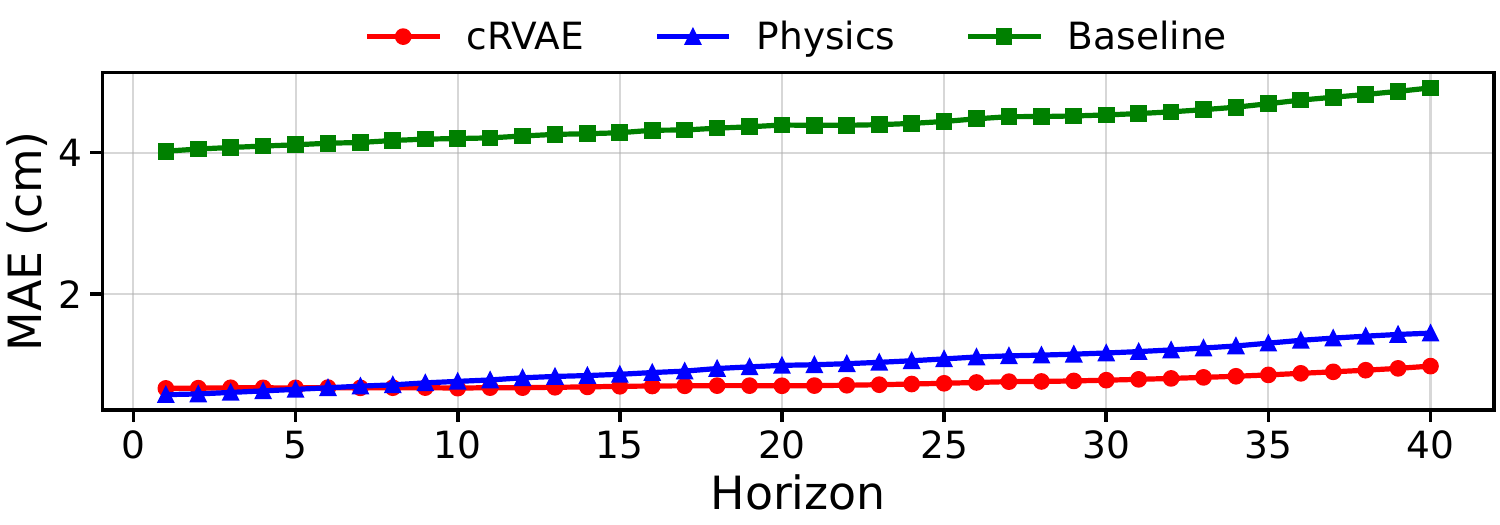}
   \caption{cRVAE prediction MAE (cm) versus prediction horizon on the fabric simulation, against a similarity-transform baseline and an XPBD cloth reference.}
    \label{fig:mse_fabric}
\end{figure}

\begin{figure}[b!]
    \centering
    \includegraphics[width=0.9\linewidth]{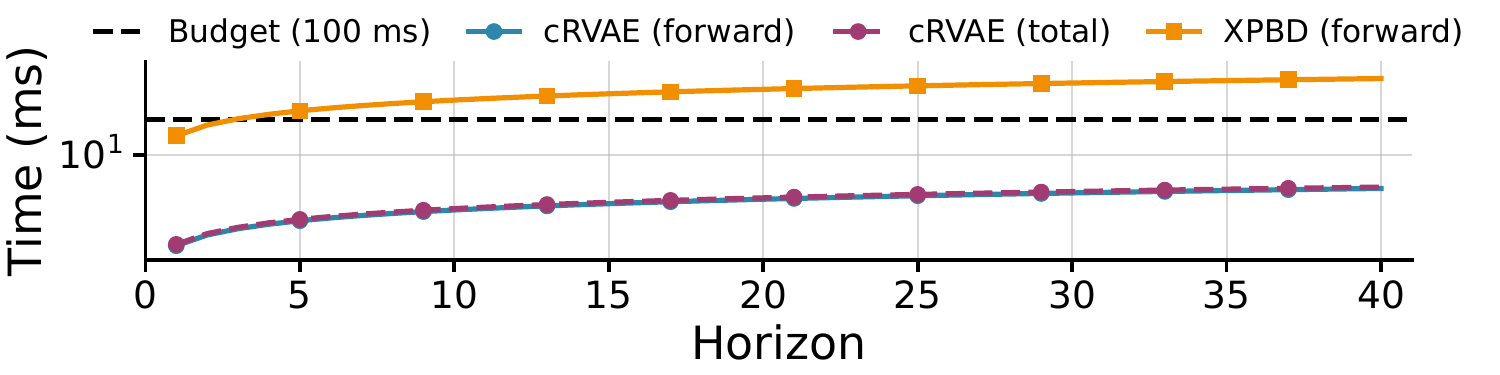}
\caption{Per-horizon compute time on the fabric ($15\times15$ mesh, log scale) versus horizon $h$. ``Total'' is shown for cRVAE only and includes derivative computation.}
    \label{fig:timing_fabric}
\end{figure}

The cRVAE is trained on fabric dynamics generated by a mesh-based Hooke's law simulator \cite{mesh-hook-law-sim-fabric}. The fabric is modeled as a $15 \times 15$ node mesh spanning a $1\,\text{m} \times 1\,\text{m}$ area. Nodes 0 and 210 act as the actuated nodes, with the input $f(k) \in \mathbb{R}^6$ given by their velocities in the $x$, $y$, and $z$ directions at a time step of $\Delta t = 0.1$\,s, while nodes 14 and 224 serve as the remaining corner nodes, giving $\mathcal{C}_k = \{x^{0}_k, x^{210}_k, x^{14}_k, x^{224}_k\}$. The training data consists of 29 trajectories of 12 seconds each, covering both joint and independent movements of the two actuated nodes.
As shown in Fig.~\ref{fig:mse_fabric}, 
using the same MAE metric defined in Section~\ref{sec:rope_sim}, averaged over all 225 nodes,
the physics reference is the XPBD counterpart for cloth, with stretching, shear, and bending constraints. cRVAE reaches $0.66$~cm at $h=1$ and $0.98$~cm at $h=40$, against $0.57$ and $1.44$~cm for the physics reference and $4.02$ and $4.92$~cm for the baseline. At short horizons cRVAE and the physics reference are within $0.1$~cm of each other, but their error curves diverge as the horizon grows, with cRVAE $0.46$~cm below the physics reference at $h=40$. Both remain below the baseline, which stays above $4$~cm across all horizons. While cRVAE and the physics reference are comparable in accuracy, they differ in compute cost. Fig.~\ref{fig:timing_fabric} reports the compute time on the fabric mesh. Per forward pass, XPBD is roughly $1550\times$ slower than cRVAE and exceeds the $100$\,ms budget at $h=3$, whereas cRVAE stays at $1.5$\,ms at $h=40$ even including the controller derivatives. The gap is wider than on the rope because XPBD's cost grows with the number of nodes and constraints. XPBD's total time is omitted, as its Jacobian exceeds memory at this state dimension. cRVAE thus matches a parameter-identified physics model in accuracy while estimating the full state from only sparse corner nodes and remaining fast enough for in-the-loop planning.

\subsubsection*{b) Comparison}

The proposed method is compared against \cite{comparison-burak}, 
which proposes a reactive control barrier function (CBF)-based 
quadratic programming (QP) controller that computes the control 
input from the current distance to the obstacle, without 
predicting future object states over a horizon; we refer to 
this method as \textit{CBF-QP} throughout this paper.  The method is evaluated in a fabric manipulation task where nodes 14 and 224 act as the leaders and nodes 0 and 210 as the followers, which should maintain a desired offset of $(0, -0.2, 0.4)$\,m while avoiding obstacles.
The leaders move forward in the $y$ direction at a mean speed of $0.02$\,m/s with a standard deviation of $10\%$ of the mean. Two obstacles are placed on top of each other along the path: the lower obstacle is a cube with side length $0.9$\,m centered at $(0.0, 0.5, 0.45)$\,m, and the upper obstacle is a box with dimensions $3\,\text{m} \times 3\,\text{m} \times 0.5$\,m centered at $(0.0, 0.5, c_2)$\,m, where $c_2$ varies across four scenarios to narrow the feasible path.
The lower cube occupies $z \in [0, 0.9]$\,m and the upper box
$z \in [c_2 - 0.25,\, c_2 + 0.25]$\,m, so the free vertical passage between
them is $g = c_2 - 1.15$\,m. Each scenario is repeated 10 times using an MPPI solver. 

For free gaps $g \geq 2.1$\,m and above, both methods complete most runs; in this wide-gap regime the clearance is governed by the lower obstacle rather than the gap, and cRVAE keeps a larger obstacle margin than CBF-QP ($0.23$\,m vs.\ $0.029$\,m) at the cost of higher tracking error ($0.98$\,m vs.\ $0.43$\,m), widening the desired offset to preserve clearance. As the gap narrows, CBF-QP's success rate drops to $0\%$ for $g \leq 1.6$\,m, while cRVAE completes all runs by stretching the fabric through the gap.

\begin{table}[t!]
    \centering
    \vspace*{5pt}
    \caption{Task Completion -- Fabric Scenarios}
    \label{tab:safety_fabric}
    \resizebox{\linewidth}{!}{
    \begin{tabular}{lccccc}
        \hline
        \textbf{Method} 
        & \textbf{Compl.}
        & \textbf{Free gap $g$ (m)}
        & \textbf{Clearance}
        & \textbf{Max Err.\ P1}
        & \textbf{Max Err.\ P2} \\
            \hline
        CBF-QP \cite{comparison-burak}     & $100\%$ & $3.1$ & $0.029 \pm 0.016$ & $0.43$ & $0.44$ \\
         cRVAE (Ours)   & $100\%$ & $3.1$ & $0.23 \pm 0.005$  & $0.98$ & $0.52$ \\
        \hline
        CBF-QP \cite{comparison-burak}    & $80\%$  & $2.1$ & $0.0 \pm 0.0$     & $0.90$ & $0.40$ \\
         cRVAE (Ours)   & $100\%$ & $2.1$ & $0.23 \pm 0.005$  & $0.98$ & $0.52$ \\
        \hline
        CBF-QP \cite{comparison-burak}    & $0\%$   & $1.6$ & N/A                & N/A    & N/A \\
    cRVAE (Ours)  & $100\%$ & $1.6$ & $0.23 \pm 0.004$  & $0.98$ & $0.59$ \\
        \hline
        CBF-QP \cite{comparison-burak}     & $0\%$   & $1.1$ & N/A                & N/A    & N/A \\
         cRVAE (Ours)   & $100\%$ & $1.1$ & $0.20 \pm 0.01$   & $1.14$ & $1.17$ \\
        \hline
    \end{tabular}}
\end{table}

\subsection{Real-World Experiments}
\label{sec:realworld_exp}
We validate the proposed method on a real rope manipulation 
task using a Unitree Go2 robot. We first describe the 
experimental setup and data collection, followed by simulator 
calibration to reduce the sim-to-real gap and cRVAE training 
to reduce the real-to-sim gap, and finally report the 
real-world control results.

\subsubsection*{a) Experimental Setup}
The real-world experiments are conducted using a Unitree Go2 
quadruped robot equipped with a Unitree D1 manipulator, which 
grasps one end of the rope and acts as the actuated node. The 
other end is held by an SO-101 manipulator, which acts as the 
leader. The rope has a length of $180$\,cm.  For capturing the rope state, six motion capture sensors are 
placed uniformly along the rope, providing the $x$, $y$, $z$ 
position of each sensor node. Nodes 0 and 5, corresponding to 
the two rope endpoints, are used as the corner nodes by the 
cRVAE during inference. The six-marker configuration is used for data collection and
calibration only; during the control runs the markers are
reallocated to the two rope corners, the obstacle, and the robot
base, so no ground truth is available for the interior nodes
during control. Eight real-world datasets are collected using the motion capture system, including random motions. 

\begin{figure*}[t]
    \centering
    \begin{subfigure}[t]{0.28\linewidth}
        \centering
        \includegraphics[width=\linewidth]{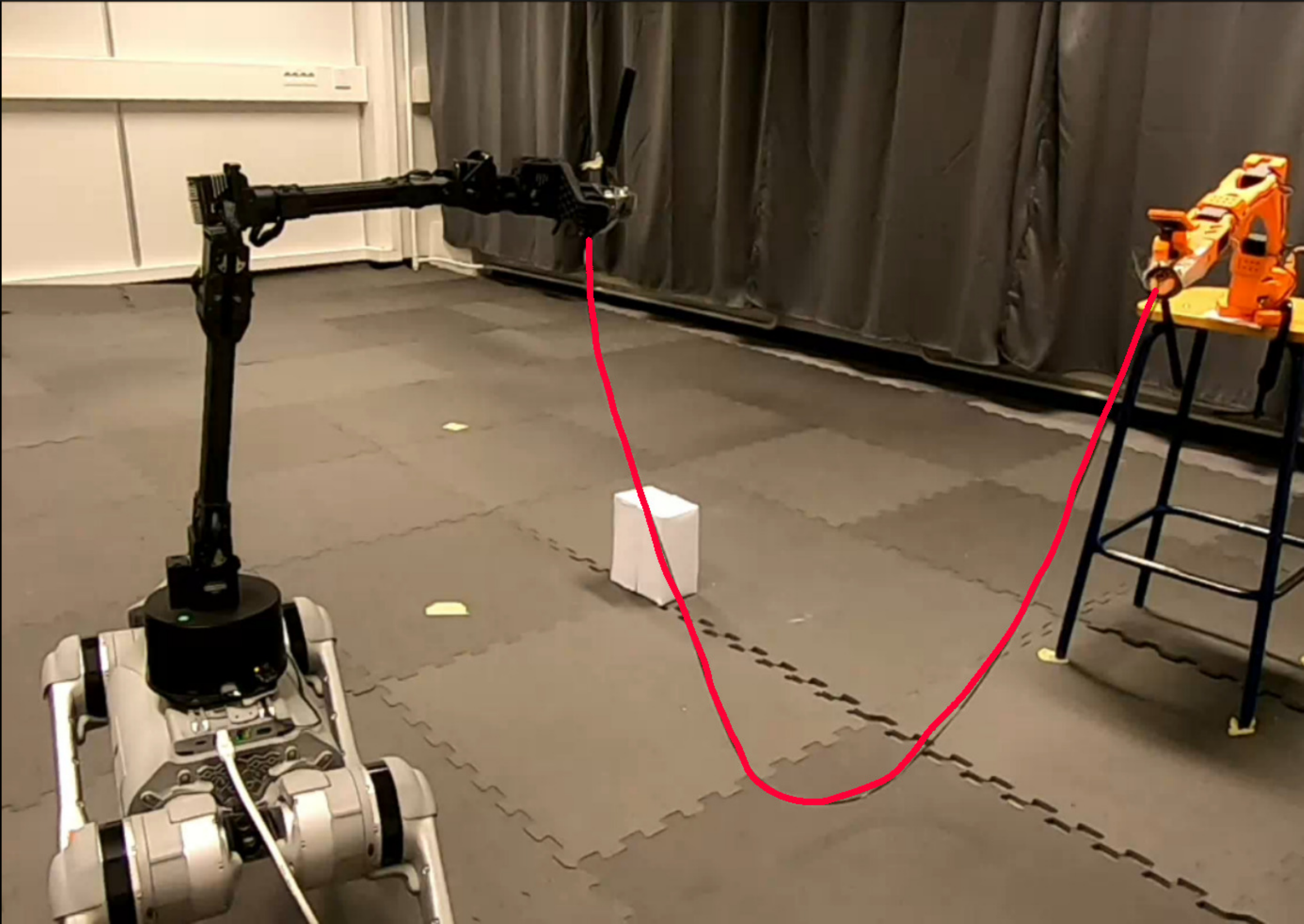}
        \caption{$t=0$ s.}
    \end{subfigure}%
    \hfill
    \begin{subfigure}[t]{0.28\linewidth}
        \centering
        \includegraphics[width=\linewidth]{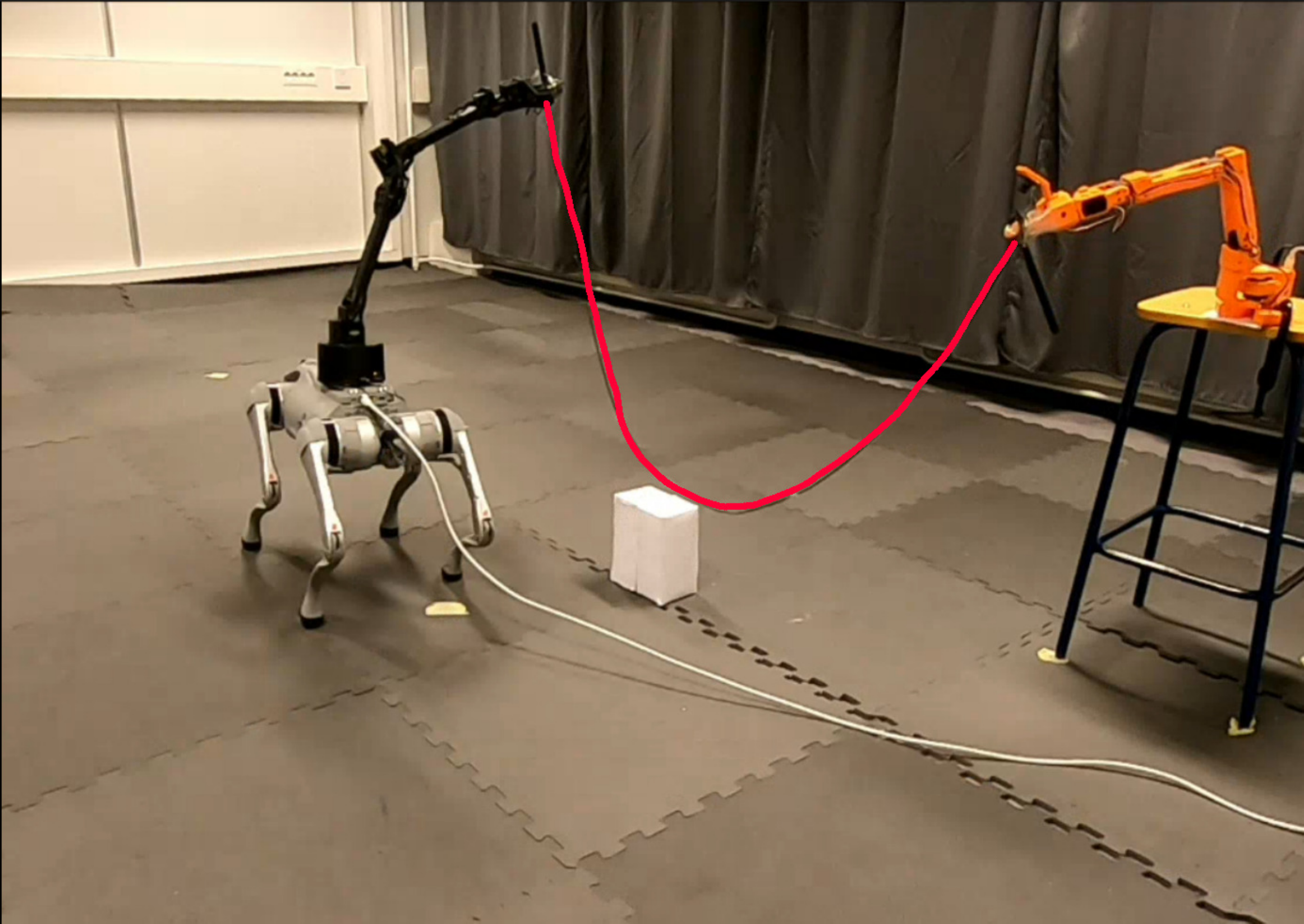}
        \caption{$t=12$ s.}
    \end{subfigure}%
    \hfill
    \begin{subfigure}[t]{0.28\linewidth}
        \centering
        \includegraphics[width=\linewidth]{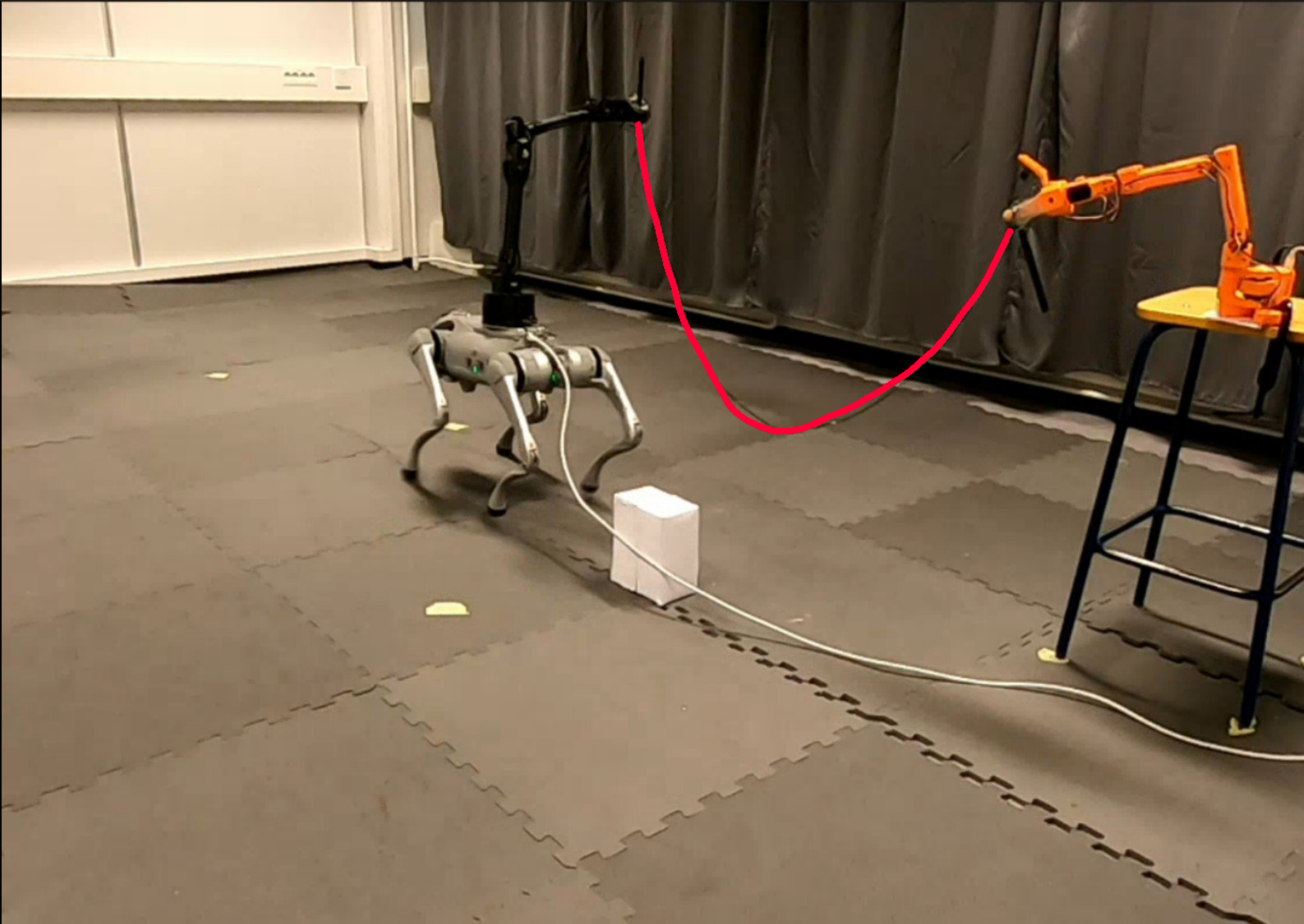}
        \caption{$t=19$ s.}
    \end{subfigure}
\caption{Real-world obstacle avoidance, shown as snapshots
from a fixed camera with the rope highlighted in red for
visibility. At $t=0$ s (a),
the rope approaches the box at low height. At $t=12$ s (b),
the manipulator lifts the rope to clear the box. At $t=19$ s
(c), the manipulator returns to a lower height after
clearing the obstacle.}
    \label{fig:realworld_photos}
\end{figure*}

\subsubsection*{b) Simulator Calibration}
To reduce the sim-to-real gap, the XPBD simulator's physical parameters, including the bending and stretching stiffness and damping coefficients, the rope mass and rest length, and the continuous positions of the six motion capture sensors along the rope, are calibrated by minimizing the difference between the simulated and real sensor positions over a single real-world dataset. The optimization is solved using differential evolution to find a global solution. A single real-world dataset is used for calibration, since the underlying optimization problem is computationally expensive. Using the optimized parameters, additional simulated data is generated covering circular movements and translations along the $x$, $y$, and $z$ directions.

\subsubsection*{c) cRVAE Training and Calibration}
The cRVAE is first trained on high-resolution simulated data 
generated using the calibrated XPBD simulator, where the rope 
is modeled with $N = 31$ nodes. To 
reduce the real-to-sim gap, the decoder is 
calibrated on the real-world data while the encoder, prior, 
and RNN weights are kept frozen. Since the encoder expects 
the full $31$-node rope state while the real-world data 
provides only 6 mocap markers, the model is run in prior 
sampling mode during calibration — bypassing the encoder 
entirely and sampling the latent variable from the prior (i.e.,
$p_\psi(z \mid \eta_k)$). The decoder is the only component 
retrained, as it directly maps RNN output to physical rope 
 and has the most direct influence on the 
real-to-sim gap.

Since the cRVAE predicts the full high-resolution rope state 
of $31$ nodes but only 6 ground truth positions are available 
from the real-world mocap sensors, no ground truth is available 
for the remaining intermediate nodes. The stretching and 
bending constraints are therefore included in the loss to 
provide gradient signal for these nodes, and the calibration 
loss is defined as:
\begin{equation}
\begin{aligned}
\mathcal{L}_{cal} = \sum_{k=1}^{H} \gamma^k
\Bigl[
&\|\tilde{P}_k - \hat{P}_k[\mathcal{M}]\|^2
+ w_c\,\|\mathcal{C}_k - \widehat{\mathcal{C}}_k\|^2  \\
&+ \lambda_s\,\mathcal{L}_{stretch}(\hat{P}_k)
+ \lambda_b\,\mathcal{L}_{bend}(\hat{P}_k)
\Bigr],
\end{aligned}
\label{eq:cal_loss}
\end{equation}
where $\tilde{P}_k \in \mathbb{R}^{18}$ are the real 
mocap marker positions, $\hat{P}_k \in \mathbb{R}^{93}$ 
is the predicted full rope state, 
$\hat{P}_k[\mathcal{M}]$ are 
the predicted positions at the indices corresponding to the 
real-world marker locations, $\mathcal{C}_k, \widehat{\mathcal{C}}_k \in \mathbb{R}^6$ are the measured and predicted 
endpoint positions, $\lambda_s, \lambda_b \in \mathbb{R}$ are 
weighting coefficients for the stretching and bending terms, 
$w_c \in \mathbb{R}$ is the corner-node loss weight, and 
$\gamma \in (0,1]$ is a discount factor.  The stretching and bending terms, $\mathcal{L}_{stretch}$ and $\mathcal{L}_{bend}$, provide gradient signal for all 31 nodes by penalizing deviations from the rope's rest segment length between neighboring nodes and from sharp bends along the rope. 
\\The decoder is calibrated on seven of the eight collected real-world datasets, with one held out for validation.
Training the decoder with this calibration loss reduces the real-world MAE from 4.05 cm to 2.63 cm on the calibration trajectory, at the cost of increasing the simulation MAE from 1.32 cm to 4.22 cm (Table~\ref{tab:calib_validation}). On a held-out validation trajectory, MAE also
decreases after calibration, from $2.88$~cm to $1.85$~cm.
 Looking at the error per marker, before calibration the model consistently
predicts the rope higher than it really is. This bias is largest in the middle
of the rope ($3.73$~cm at node~18) and almost zero at the grasped end
($0.01$~cm at node~30), as shown in Table~\ref{tab:zbias}. Since the error is in
the middle of the rope and not at the end effector, the manipulator has to lift
more than the error itself to correct it, and this lift is bounded by the D1's
vertical workspace. Calibration reduces this mid-span offset from
$3.73$~cm to $1.09$~cm.

\begin{table}[t!]
\vspace*{5pt}
\centering
\caption{Decoder calibration MAE (cm): simulation, the real-world
calibration set, and a held-out real-world validation set.}
\label{tab:calib_validation}
\begin{tabular}{lccc}
\hline
 & Sim & Real  & Validation \\
\hline
Before calibration & 1.32 & 4.05 & 2.88 \\
After calibration  & 4.22 & 2.63 & 1.85 \\
\hline
\end{tabular}
\end{table}
\begin{table}[t]
\vspace*{5pt}
\centering
\caption{Mean signed $z$-error (cm) per marker on the real-world calibration
set. Positive values indicate the model predicts the rope above its true
position, i.e.\ it underestimates the sag.}
\label{tab:zbias}
\resizebox{\columnwidth}{!}{%
\begin{tabular}{|l|c|c|c|c|c|c|}
\hline
Node & 0 & 6 & 12 & 18 & 24 & 30 \\
\hline
Before calibration & 1.09 & 1.38 & 1.95 & \textbf{3.73} & 1.35 & 0.01 \\
\hline
After calibration  & 0.01 & $-0.52$ & $-1.03$ & $-1.09$ & $-0.47$ & $-0.01$ \\
\hline
\end{tabular}
}
\end{table}

\subsubsection*{d) Real-World Control Experiments}
\label{sec:realworld_control}
The proposed method is validated on a real-world rope 
manipulation task using two scenarios with different obstacle 
geometries. The controller plans over a horizon of 5, while new corner measurements arrive at every control step. In both scenarios, the Unitree Go2 acts as the 
follower and the SO-101 arm as the leader. The follower 
is required to maintain a desired radial offset and 
angular synchrony with the leader along a circular arc. 
The $x$ and $y$ motion of the follower is controlled 
by the Go2 base, while the $z$ motion is controlled 
by the D1 manipulator, keeping the two axes decoupled.
The problem is solved using iLQR~\cite{ilqr} in an event-based manner. The purpose of this experiment is to demonstrate feasibility on physical hardware. cRVAE runs on the real system using only the two corner nodes available from sensing, within the control budget. A direct comparison against \cite{comparison-burak} is challenging, since it requires the full rope state, which the real hardware does not provide.

Scenario 1 uses an obstacle of dimensions $0.075\,\text{m} \times 0.09\,\text{m} \times 0.18\,\text{m}$, 
while Scenario 2 uses $0.156\,\text{m} \times 0.07\,\text{m} \times 0.10\,\text{m}$. 
Each scenario is evaluated across 10 runs with different initial positions 
of the robot and obstacle. Table~\ref{tab:realworld} summarizes the results 
across both scenarios. 
Task success is determined directly from the recorded videos of each run by checking for physical contact between the rope and the obstacle; no collision was observed in any run. The computation time per control action, $65.6$ and $63.6$\,ms across the two scenarios (Table~\ref{tab:realworld}), stays within the $100$\,ms control period set by $\Delta t = 0.1$\,s.

\begin{table}[t!]
    \centering
    \caption{Real-world experiment results across two scenarios.}
    \label{tab:realworld}
    \renewcommand{\arraystretch}{1.3}
    \resizebox{\linewidth}{!}{
    \begin{tabular}{lccccc}
        \hline
        \textbf{Scenario} 
        & $\bar{e}_{xy}$ \textbf{(mm)} 
        & $\bar{d}_z$ \textbf{(mm)}
        & $t_{\text{comp}}$ \textbf{(ms)}
        & \textbf{Successful Runs} \\
        \hline
        Scenario 1 & $50.5 \pm 3.4$ & $138.6 \pm 55.9$ & $65.57 \pm 9.94$ & $10/10$ \\
        Scenario 2 & $52.8 \pm 4.7$ & $78.1 \pm 6.0$ & $63.58 \pm 6.38$ & $10/10$ \\
        \hline
    \end{tabular}}
    \vspace{0.5pt}
\footnotesize{$\bar{e}_{xy}$: mean XY leader--follower error; $\bar{d}_z$: mean manipulator lift; $t_{\text{comp}}$: computation time per control action; Successful Runs: runs with no observed rope--obstacle contact, out of 10 runs per scenario. Values reported as mean $\pm$ std across 10 runs per scenario.}
\end{table}

\section{CONCLUSION}
This work presented cRVAE, a conditional recurrent variational
autoencoder for modeling deformable object dynamics from partial
node observations, integrated into a receding-horizon optimal
control framework for manipulation. In simulation, at horizon $h=40$ cRVAE reached $2.51$~cm on the rope and
$0.98$~cm on the fabric, against $3.28$ and $1.44$~cm for a
parameter-identified XPBD reference and $5.66$ and $4.92$~cm for a geometric
baseline. It estimates the full state from two observed nodes on the rope and four on the fabric. A rollout including derivatives takes $\approx 9$~ms on the rope
against $3.2$~s for XPBD, and $1.5$~ms on the fabric, where XPBD is
$\approx 1550\times$ slower per forward pass. On real hardware, cRVAE completed
10/10 rope runs on a Unitree Go2 without observed obstacle contact. 

Future work will extend cRVAE to model contact, enabling
manipulation tasks in which the object interacts with its environment.
A second direction is to extend the measurement feedback with vision,
using it to obtain interior measurements that the two corner nodes
cannot supply and thereby improve the prediction accuracy of the
unmeasured nodes.

\end{document}